\documentclass[10pt,twocolumn,letterpaper]{article}

\usepackage{cvpr}              
\usepackage[accsupp]{axessibility} 

\usepackage[dvipsnames]{xcolor}

\definecolor{cvprblue}{rgb}{0.21,0.49,0.74}
\usepackage[pagebackref,breaklinks,colorlinks,citecolor=cvprblue]{hyperref}

\usepackage{multirow}

\def\paperID{14599} 
\def\confName{CVPR}
\def\confYear{2024}

\title{Optimal Transport Aggregation for Visual Place Recognition}

\author{Sergio Izquierdo \qquad Javier Civera\\
I3A, University of Zaragoza, Spain\\
{\tt\small \{izquierdo, jcivera\}@unizar.es}
}

\begin{document}
\maketitle
\begin{abstract}
The task of Visual Place Recognition (VPR) aims to match a query image against references from an extensive database of images from different places, relying solely on visual cues.
State-of-the-art pipelines focus on the aggregation of features extracted from a deep backbone, in order to form a global descriptor for each image.
In this context, we introduce SALAD (\underline{S}inkhorn \underline{A}lgorithm for \underline{L}ocally \underline{A}ggregated \underline{D}escriptors), which reformulates NetVLAD's soft-assignment of local features to clusters as an optimal transport problem.
In SALAD, we consider both feature-to-cluster and cluster-to-feature relations and we also introduce a `dustbin' cluster, designed to selectively discard features deemed non-informative, enhancing the overall descriptor quality.
Additionally, we leverage and fine-tune DINOv2 as a backbone, which provides enhanced description power for the local features, and dramatically reduces the required training time.
As a result, our single-stage method not only surpasses single-stage baselines in public VPR datasets, but also surpasses two-stage methods that add a re-ranking with significantly higher cost. Code and models are available at \href{https://github.com/serizba/salad}{https://github.com/serizba/salad}.
\end{abstract}

\section{Introduction}
\label{sec:intro}

Recognizing a place solely from images becomes a challenging task when scenes undergo substantial changes in their structure or appearance. Such capability is referred to in the scientific and technical literature as visual place recognition (and by its acronym VPR), and is essential for agents to navigate and understand their surroundings autonomously in a wide array of applications, such as robotics~\cite{chen2017only,chen2017deep,chen2018learning,khaliq2019holistic, hausler2019multi} or augmented reality~\cite{garg2021your}. Specifically, it is present in simultaneous localization and mapping \cite{cadena2016past,campos2021orb} and absolute pose estimation \cite{irschara2009structure,pion2020benchmarking} pipelines. 

\begin{figure}[t!]
  \centering
   \includegraphics[width=0.99\linewidth]{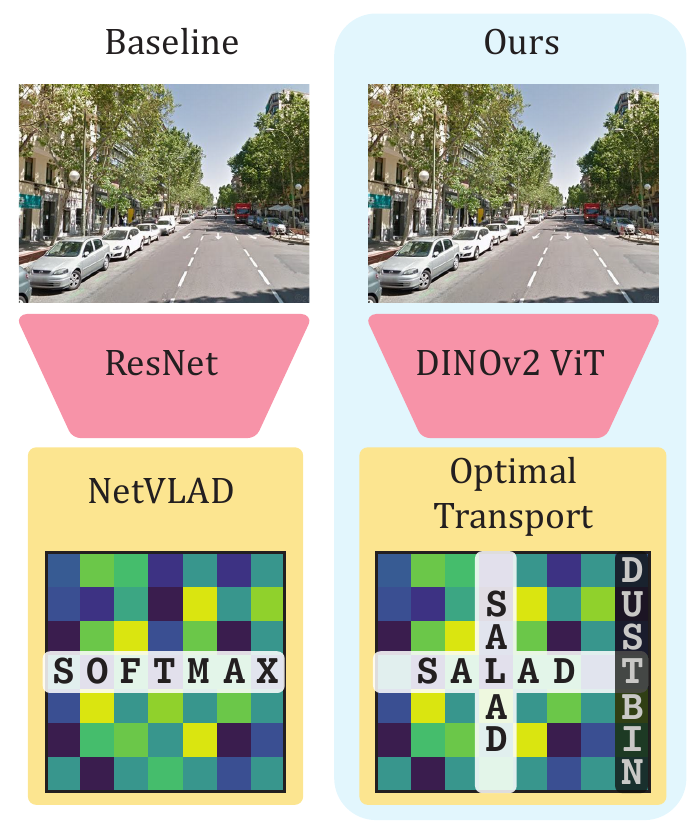}
    \caption{\textbf{Illustration of a VPR baseline (left) and our contribution (right).} The left column outlines a typical VPR baseline, a ResNet backbone followed by NetVLAD aggregation~\cite{arandjelovic2016netvlad}. On the right column, we replace ResNet with a partially fine-tuned DINOv2~\cite{oquab2023dinov2} backbone, and incorporate SALAD, our novel optimal transport aggregation using the Sinkhorn Algorithm. Our model achieves unprecedented state-of-the-art results on common VPR benchmarks.}
   \label{fig:teaser}
\end{figure}\vspace{0.5em}

In practice, VPR is framed as an image retrieval problem, wherein typically a query image serves as the input and the goal is to obtain an ordered list of top-k matches against a pre-existing database of geo-localized reference images.
Images are represented as an aggregation of appearance pattern descriptors, which are subsequently compared via nearest neighbour.
The effectiveness of this matching relies on generating discriminative per-image descriptors that exhibit robust performance even for challenging variations such as fluctuating illumination, structural transformations, temporal changes, weather and seasonal shifts.
Most recent research on VPR have thus focused on the two key components of this general pipeline, namely the deep neural backbones for feature extraction and methods for aggregating such features.

For years, ResNet-based neural networks have been the predominant backbones for feature extraction~\cite{arandjelovic2016netvlad,radenovic2018fine,hausler2021patch}. Recently, given the success of Vision Transformer (ViT) for different computer vision tasks~\cite{dosovitskiy2020image,han2022survey,lee2022mpvit,liu2022swin}, some methods have introduced ViT in the field of VPR~\cite{wang2022transvpr, zhu2023r2former}. AnyLoc~\cite{keetha2023anyloc} proposed to leverage foundation models, using DINOv2~\cite{oquab2023dinov2} as a feature extractor for VPR. However, AnyLoc uses DINOv2 `as is', while we show in this paper that fine-tuning the model for VPR brings a significant increase in performance.

Regarding aggregation, NetVLAD~\cite{arandjelovic2016netvlad}, the learned counterpart to the traditional handcrafted VLAD~\cite{jegou2010aggregating}, is among the most popular choices.
Alternative methods include pooling layers like GeM~\cite{radenovic2018fine} or learned global aggregation, like the recent MixVPR~\cite{ali2023mixvpr}. In this paper, we propose optimal transport aggregation, setting a new state of the art in VPR.

As a summary, in this work, we present a single-stage approach to VPR that obtains state-of-the-art results in the most common benchmarks. To achieve this, we present two key contributions:

\begin{itemize}
\item First, we propose SALAD (\underline{S}inkhorn \underline{A}lgorithm for \underline{L}ocally \underline{A}ggregated \underline{D}escriptors), a reformulation of the feature-to-cluster assignment problem through the lens of optimal transport, allowing more effective distribution of local features into the global descriptor bins. To further improve the discriminative power of the aggregated descriptor, we let the network discard uninformative features by introducing a `dustbin' mechanism.

\item Secondly, we integrate the representational power of foundation models into VPR, using DINOv2 as the backbone for feature extraction. Unlike previous approaches that utilized DINOv2 in its pre-trained form, our method involves fine-tuning the model specifically for the task. This fine-tuning process converges extremely fast, in just four epochs, and allows DINOv2 to capture more relevant and distinctive features pertinent to place recognition tasks. 
\end{itemize}

The fusion of these two novel components results in DINOv2 SALAD, which can be efficiently trained in less than one hour and sets unprecedented recall in VPR benchmarks, with 75.0\% Recall@1 in MSLS Challenge and 76.0\% in Nordland. All of this with a single-stage pipeline, without requiring expensive post-processing steps and with an inference speed of less than 3 ms per image.

\section{Related Work}
\label{sec:related}

The significant research efforts on VPR have been exhaustively compiled in a number of surveys and tutorials over the years~\cite{lowry2015visual,zhang2021visual,masone2021survey,garg2021your,Schubert2023vpr}. Current research addresses a wide variety of topics, such as novel loss functions~\cite{berton2022rethinking,leyva2023data}, image sequences~\cite{warburg2020mapillary,garg2022seqmatchnet}, extreme viewpoint changes~\cite{lin2015learning} or text features~\cite{hong2019textplace}. In this section, we focus on work related to feature extraction and aggregation, as there lie our contributions. 

Early approaches to VPR used either aggregations of handcrafted local features~\cite{cummins2008fab,jegou2010aggregating,arandjelovic2013all} or global descriptors~\cite{sunderhauf2011brief,murillo2012localization}. In both cases, geometric~\cite{galvez2012bags} and temporal~\cite{galvez2012bags,milford2012seqslam} consistency was sometimes enforced for enhanced performance. With the emergence of deep neural networks, features pre-trained for recognition tasks, without fine-tuning, showed a significant performance boost over handcrafted ones~\cite{sunderhauf2015performance}. However, training or fine-tuning specifically for VPR tasks using contrastive or triplet losses~\cite{musgrave2020metric} offers an additional improvement and is standard nowadays.

NetVLAD~\cite{arandjelovic2016netvlad} is the most popular architecture explicitly designed for VPR, mimicking the VLAD aggregation~\cite{jegou2010aggregating} but jointly learning from data both convolutional features and cluster centroids. \citet{radenovic2018fine} proposed the Generalized Mean Pooling (GeM) to aggregate feature activations, also a popular baseline due to its simplicity and competitive performance. In addition to these, several other alternatives have been proposed in the literature. For example, \citet{teichmann2019detect} aggregates regions instead of local features. Recently, MixVPR~\cite{ali2023mixvpr} has presented the best results in the literature by combining deep features with a MLP layer. 

A notable trend in VPR has been the adoption of a two-stage approach to enhance retrieval accuracy~\cite{taira2018inloc,sarlin2019coarse,cao2020unifying,hausler2021patch,shao2023global,zhu2023r2former}. After a first stage with any of the methods presented in the previous paragraph, the top retrieved candidates are re-ranked attending to the un-aggregated local features, either assessing the geometric consistency to the query image or predicting their similarity. This re-ranking stage adds a considerable overhead, which is why it is only applied to a few candidates, but generally improves the performance. Re-ranking is out of the scope of our research but, notably, we outperform all baselines that employ re-ranking even if our model does not include such stage (and hence it is substantially faster).

Optimal transport has found a significant number of applications in graphics and computer vision~\cite{bonneel2023survey}. Specifically, related to our research, it has been used for image retrieval~\cite{pele2009fast}, image matching~\cite{xing2022differentiable} and feature matching~\cite{sarlin2020superglue,sun2021loftr}. Recently, \citet{zhang2022beyond} used optimal transport at the re-ranking stage in a retrieval pipeline. However, ours is the first work that proposes the formulation of local feature aggregation from an optimal transport perspective.

\begin{figure*}[t]
  \centering
   \includegraphics[width=0.99\textwidth]{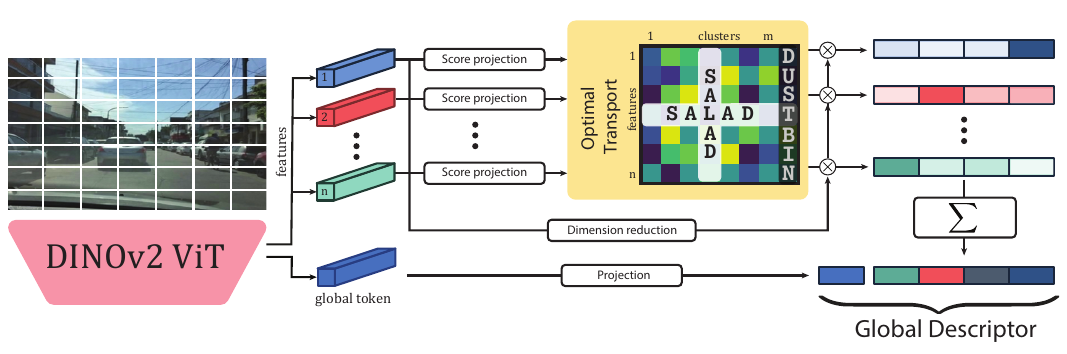}
   \caption{\textbf{Overview of our method}. First, the DINOv2 backbone extracts local features and a global token from an input image. Then, a small MLP, score projection, computes a score matrix for feature-to-cluster and dustbin relationships. The optimal transport module uses the Sinkhorn algorithm to transform this matrix into an assignment, and subsequently, dimensionality-reduced features are aggregated into the final descriptor based on this assignment and concatenated with the global token.}
   \label{fig:method}
\end{figure*}

\section{Method}
\label{sec:method}

DINOv2 SALAD is based on NetVLAD, but we propose to use and fine-tune the DINOv2 backbone (\cref{subsec:extraction}) and propose a novel module (SALAD) for the assignment (\cref{subsec:assignment}) and aggregation (\cref{subsec:aggregation}) of features.

\subsection{Local Feature Extraction}
\label{subsec:extraction}

Effective local feature extraction lies in striking a balance: features must be robust enough to withstand substantial changes in appearance, such as those between seasons or from day to night, yet they should retain sufficient information on local structure to enable accurate matching.

Inspired by the success of ViT architectures in many computer vision tasks and by AnyLoc~\cite{keetha2023anyloc}, that leverages the exceptional representational capabilities of foundation models~\cite{bommasani2021opportunities}, we adopt DINOv2~\cite{oquab2023dinov2} as our backbone. However, differently from AnyLoc, we use a supervised pipeline and include the backbone in the end-to-end training for the specific task, yielding improved performance.

DINOv2 adopts a ViT architecture that initially divides an input image $\mathbf{I}\in\mathbb{R}^{h\times w\times c}$ into $p\times p \times c$ patches, with $p=14$. These patches are sequentially projected with transformer blocks, resulting in the output tokens $\{\mathbf{t}_1, \dots, \mathbf{t}_{n}, \mathbf{t}_{n+1}\}, \mathbf{t}_i\in\mathbb{R}^{d}$, where $n=hw/p^2$ is the number of input patches and there is an additional global token $\mathbf{t}_{n+1}$ that aggregates class information. Although the DINOv2's authors reported that fine-tuning the model only brings dim improvements, we found that at least for VPR there are substantial gains in selectively unfreezing and training the last blocks of the encoder.

\subsection{Assignment}
\label{subsec:assignment}

In NetVLAD, a global descriptor is formed by assigning a set of features to a set of clusters, $\{C_1, \dots, C_j, \dots, C_m\}$, and then aggregating all features that belong to each cluster. For the assignment, NetVLAD computes a score matrix $\mathbf{S} \in \mathbb{R}_{>0}^{n \times m}$, where the element in its $i^{\text{th}}$ row and $j^{\text{th}}$ column, $s_{i,j} \in \mathbb{R}_{>0}$, represents the cost of assigning a feature to a cluster $C_j$. In other words, $\mathbf{S}$ quantifies the affinity of each feature to each clusters. While SALAD draws inspiration from NetVLAD, we identify several crucial aspects in their assignment and propose alternatives to address these.

\textbf{Reduce assignment priors.} When building the score matrix $\textbf{S}$, NetVLAD introduces certain priors. Specifically, it initializes the linear layer that computes $\textbf{S}$ with centroids derived from k-means. While this may accelerate the training, it introduces inductive bias and potentially makes the model more susceptible to local minima. In contrast, we propose to learn each row $\textbf{s}_i$ of the score matrix from scratch with two fully connected layers initialized randomly:
\begin{equation}
\textbf{s}_i =  \mathbf{W}_{s_2}(\sigma(\mathbf{W}_{s_1}(\textbf{t}_{i}) + \mathbf{b}_{s_1})) + \mathbf{b}_{s_2}
\end{equation}
where $\mathbf{W}_{s_1}$, $\mathbf{W}_{s_2}$ and $\mathbf{b}_{s_1}$, $\mathbf{b}_{s_2}$ are the weights and biases of the layers, and $\sigma$ is a non-linear activation function.

\textbf{Discard uninformative features.} Some features, such as those representing the sky, might contain negligible information for VPR. NetVLAD does not account for this, and the contribution of all features is preserved in the final descriptor. Contrary, we follow recent works on keypoint matching and introduce a `dustbin' where non-informative features are assigned to. For that, we augment the score matrix, from $\textbf{S}$ to $\bar{\textbf{S}} = \left[ \textbf{S}, \ \bar{\textbf{s}}_{i, m+1}\right] \in \mathbb{R}_{>0}^{n \times m+1}$, by appending the column $\bar{\textbf{s}}_{i, m+1}$ representing the feature-to-dustbin relation. As in SuperGlue~\cite{sarlin2020superglue}, this score is modeled with a single learnable parameter $z \in \mathbb{R}$:
\begin{equation}
\bar{\textbf{s}}_{i, m+1} = z\textbf{1}_{n}
\end{equation}
\noindent being $\textbf{1}_{n} = \left[1, \dots, 1\right]^\top \in \mathbb{R}^n$ a $n$-dimensional vector of ones.

\textbf{Optimal assignment.} The original NetVLAD assignment computes a per-row softmax over $\mathbf{S}$ to obtain the distribution of each feature's mass across the clusters. However, this approach only considers the feature-to-cluster relationship and overlooks the reverse --the cluster-to-feature relation. For this reason, we reformulate the assignment as an optimal transport problem where the features' mass, $\boldsymbol {\mu}=\mathbf{1}_n$, must be effectively distributed among the clusters or the `dustbin', $\boldsymbol{\kappa}=[\mathbf{1}^\top_m, n-m]^\top$. We follow SuperGlue~\cite{sarlin2020superglue} and use the Sinkhorn Algorithm~\cite{cuturi2013sinkhorn,sinkhorn1967concerning} to obtain the assignment $\bar{\textbf{P}} \in \mathbb{R}^{n \times \left(m+1\right)}$ such that
\begin{equation}
    \bar{\textbf{P}}\textbf{1}_{m+1} = \boldsymbol {\mu}\quad\text{and}\quad \bar{\textbf{P}}^\top\textbf{1}_{n} = \boldsymbol {\kappa}\text{.}
\end{equation}
This algorithm finds the optimal transport assignment between distributions $\boldsymbol{\mu}$ and $\boldsymbol{\kappa}$ iteratively normalizing rows and columns from $\exp\left(\bar{\textbf{S}}\right)$. Finally, we drop the dustbin column to obtain the assignment $\textbf{P} = \left[ \textbf{p}_{\ast,1}, \dots, \textbf{p}_{\ast,m} \right]$, where $\textbf{p}_{\ast,j}$ stands for the $j^{\text{th}}$ column of $\textbf{P}$.

\subsection{Aggregation}
\label{subsec:aggregation}

Once the feature assignment in our SALAD framework is computed as detailed in \cref{subsec:assignment}, we focus on the aggregation of these assigned features to form the final global descriptor. The aggregation process in NetVLAD involves combining all features assigned to each cluster $C_j$. However, we introduce three variations:

\textbf{Dimensionality reduction.} To efficiently manage the final descriptor size, we first reduce the dimensionality of the tokens from $\mathbb{R}^d$ to $\mathbb{R}^l$. This is achieved by processing the features through two fully connected layers, precisely adjusting the size of the feature vectors while retaining the essential information from the task. 
\begin{equation}
\textbf{f}_i = \mathbf{W}_{f_2}(\sigma(\mathbf{W}_{f_1}(\textbf{t}_i) + \mathbf{b}_{f_1})) + \mathbf{b}_{f_1}
\end{equation}

\textbf{Aggregation.} Based on the assignment matrix derived using the Sinkhorn Algorithm, each feature is aggregated into its assigned cluster. Differently from NetVLAD, we do not subtract the centroids to get the residuals. We directly aggregate these features with a summation, reducing the incorporated priors about the aggregation. Viewing the resulting VLAD vector as a matrix $\textbf{V} \in \mathbb{R}^{m\times l}$, each element $V_{j,k} \in \mathbb{R}$ is computed as follows:
\begin{equation}
V_{j,k} = \sum_{i=1}^{n}{P_{i, k} \cdot f_{i,k}}
\end{equation}
where $f_{i,k}$ corresponds to the $k^{\text{th}}$ dimension of $\textbf{f}_i$, with $k \in \{1, \dots, l \}$.

\textbf{Global token.} To include global information about the scene not easily incorporated into local features, we also incorporate a scene descriptor $g$ computed as:
\begin{equation}
\textbf{g} = \mathbf{W}_{g_2}(\sigma(\mathbf{W}_{g_1}(\textbf{t}_{n+1}) + \mathbf{b}_{g_1})) + \mathbf{b}_{g_1}
\end{equation}
where $\textbf{t}_{n+1}$ is the global token from DINOv2. We then concatenate $\textbf{g}$ with $\textbf{V}$ flattened. Following NetVLAD, we do an L2 intra-normalization and an entire L2 normalization of this vector, which yields the final global descriptor.

\section{Experiments}
\label{sec:experiments}

\begin{table*}[!htbp]
    \centering
    \resizebox{\linewidth}{!}{
    \begin{tabular}{l c c c c c c c c c c c c c c c c}
    \hline

    \hline
    \multirow{2}{*}{Method} &  && \multicolumn{2}{c}{MSLS Challenge}&&\multicolumn{2}{c}{MSLS Val}&&\multicolumn{2}{c}{NordLand}&&\multicolumn{2}{c}{Pitts250k-test}&&\multicolumn{2}{c}{SPED}\\
    \cline{4-5} \cline{7-8} \cline{10-11} \cline{13-14} \cline{16-17}
    &  Desc. size & Latency (ms) & R@1 & R@5 && R@1 & R@5 && R@1 & R@5 && R@1 & R@5 && R@1 & R@5\\
    \hline
        NetVLAD~\cite{arandjelovic2016netvlad} & 32768 & 1.41 & 35.1 & 47.4 && 82.6 & 89.6 && 32.6 & 47.1 && 90.5 & 96.2 && 78.7 & 88.3\\
        GeM~\cite{radenovic2018fine}\textdagger & 1024 & 1.14 & 49.7 & 64.2 && 78.2 & 86.6 && 21.6 & 37.3 && 87.0 & 94.4 && 66.7 & 83.4\\
        Conv-AP~\cite{ali2022gsv} & 8192 & 1.22 & 54.2 & 66.6 && 83.1 & 90.3 && 42.7 & 58.9 && 92.9 & 97.7 && 79.2 & 88.6\\
        CosPlace~\cite{berton2022rethinking} & 2048 & 2.59 & 67.2 & 78.0 && 87.4 & 93.0 && 44.2 & 59.7 && 92.1 & 97.5 && 80.1 & 89.6\\
        MixVPR~\cite{ali2023mixvpr} & 4096 & 1.37 & 64.0 & 75.9 && 88.0 & 92.7 && 58.4 & 74.6 && 94.6 & 98.3 && 85.2 & 92.1\\
        EigenPlaces~\cite{berton2023eigenplaces} & 2048 & 2.65 & 67.4 & 77.1 && 89.3 & 93.7 && 54.4 & 68.8 && 94.1 & 98.0 && 69.9 & 82.9\\
    \hline
        DINOv2 SALAD & 512 + 32 & 2.33 & 70.8 & 83.6 && 89.3 & 94.9 && 61.2 & 78.9 && 93.0 & 97.4 && 88.5 & 94.7\\
        DINOv2 SALAD & 2048 + 64 & 2.35 & 73.7 & 85.9 && 90.5 & 95.4 && 70.4 & 85.7 && 94.8 & 98.3 && 89.5 & 94.9\\
        DINOv2 SALAD & 8192 + 256 & 2.41 & \textbf{75.0} & \textbf{88.8} && \textbf{92.2} & \textbf{96.4} && \textbf{76.0} & \textbf{89.2} && \textbf{95.1} & \textbf{98.5} && \textbf{92.1} & \textbf{96.2}\\
    \hline

    \hline
    \end{tabular}}
    \caption{\textbf{Comparison against single-stage baselines.} We compare DINOv2 SALAD against two popular baselines~\cite{arandjelovic2016netvlad,radenovic2018fine} and the four baselines that show best results in recent literature~\cite{ali2022gsv,berton2022rethinking,ali2023mixvpr,berton2023eigenplaces}. Our slim version already obtains state-of-the-art results in all metrics. Our full model outperforms all previous results by a significant margin. Note, in particular, the large improvement in the most challenging benchmarks, MSLS Challenge and NordLand. \textdagger~We reproduced GeM results training during 80 epochs following MixVPR training pipeline.}
\label{tab:main}
\end{table*}

\begin{table*}
    \centering
    \resizebox{\linewidth}{!}{
    \begin{tabular}{l c c c c c c c c c c c c c c c c}
    \hline

    \hline
    \multirow{2}{*}{Method} & \multicolumn{2}{c}{Desc. size} && \multirow{2}{*}{Memory (GB)} && \multicolumn{2}{c}{Latency (ms)} &&  \multicolumn{3}{c}{MSLS Challenge} && \multicolumn{3}{c}{MSLS Val}\\
    \cline{2-3} \cline{7-8} \cline{10-12} \cline{14-16}
    &  Global & Local && && Retrieval & Reranking && R@1 & R@5 & R@10 && R@1 & R@5 & R@10\\
    \hline
        Patch-NetVLAD~\cite{hausler2021patch} & $4096$ & $2826\times 4096$ &&  908.30  &&9.55 & 8377.17 && 48.1 & 57.6 & 60.5 && 79.5 & 86.2 & 87.7\\
        TransVPR~\cite{wang2022transvpr} & $\textbf{256}$ & $1200\times 256$ && 22.72 && 6.27 & 1757.70 && 63.9 & 74.0 & 77.5 && 86.8 & 91.2 & 92.4\\
        R2Former~\cite{zhu2023r2former} & $\textbf{256}$ & $500 \times 131$ && 4.7 && 8.88 & 202.37 && 73.0 & 85.9 & 88.8 && 89.7 & 95.0 & 96.2 \\
    \hline
        \textbf{DINOv2 SALAD (ours)} & $8192 + 256$ & \textbf{0.0} &&  \textbf{0.63} && \textbf{2.41} & \textbf{0.0} && \textbf{75.0} & \textbf{88.8} & \textbf{91.3} && \textbf{92.2} & \textbf{96.4} & \textbf{97.0}\\
    \hline

    \hline
    \end{tabular}
    }
    \caption{\textbf{Comparison against baselines with re-ranking.} We compare our single-stage DINOv2 SALAD with methods that perform a re-ranking stage to improve performance. Without using re-ranking, our DINOv2 SALAD outperforms all other methods while being orders of magnitude faster and more memory-efficient. Latency metrics obtained from \cite{zhu2023r2former} using a RTX A5000. Latency for DINOv2 SALAD was computed using a RTX 3090. Memory footprint is calculated on the MSLS Val dataset, which includes around $18,000$ images.}
\label{tab:reranking}
\end{table*}

\begin{table}
    \centering
    \footnotesize
    \resizebox{\linewidth}{!}{
    \begin{tabular}{l c c c}
    \hline

    \hline
    Method & Desc. size & SF-XL Test v1 & SF-XL Test v2\\
    \hline
        CosPlace~\cite{berton2022rethinking} & 2048 & 76.4 & 88.8\\
        EigenPlaces~\cite{berton2023eigenplaces} & 2048 & 84.1 & 90.8\\ \hline 
        DINOv2 SALAD & 8192 + 256 & \textbf{88.6} & \textbf{94.8}\\
    \hline

    \hline
    \end{tabular}
    }
    \vspace{-0.5em}
    \caption{\textbf{Results on SF-XL. (R@1)} Our DINOv2 SALAD achieves unprecedented results on SF-XL despite never seeing any single image of San Francisco during VPR finetuning.}
    \vspace{-0.5em}
\label{tab:sfxl}
\end{table}

To rigorously evaluate the effectiveness of our proposed contributions, we conducted exhaustive experiments following standard evaluation protocols.

\subsection{Implementation Details}
\label{subsec:implementation}

We ground our training and evaluation setups on the publicly provided framework by MixVPR\footnote{\url{https://github.com/amaralibey/MixVPR}}. 

For the \textbf{architecture}, we opt for a pretrained \mbox{DINOv2-B} backbone, targeting a balance between computational efficiency and representational capacity. We only fine-tune the final 4 layers of the encoder, which significantly enhances the performance without markedly increasing training time. For the fully connected layers, the weights of the hidden layers $\mathbf{W}_{s_1}$, $\mathbf{W}_{f_1}$ and $\mathbf{W}_{g_1}$ have $512$ neurons and use ReLU for the activation function $\sigma$. 
To optimize feature handling, we employ a dimensionality reduction, compressing feature token dimensions from $d=768$ to $l=128$, and the global to 256. We use $m=64$ clusters, resulting in a global descriptor of size $128 \times 64 + 256$. We also report results with smaller descriptors, with size $512+32$ ($m=15,\ l=32$), and $2048+64$ ($m=32,\ l=64$).

We \textbf{train} on GSV-Cities~\cite{ali2022gsv}, a large dataset of urban locations collected from Google Street View. Given the impressive representation power of DINOv2, our pipeline achieves training convergence within just 4 epochs. Using a batch size of 60 places, each represented by 4 images, the training is completed in 30 minutes on a single NVIDIA RTX 3090. We use the multi-similarity loss~\cite{wang2019multi} and AdamW~\cite{loshchilov2017decoupled} for the optimization, with an initial learning rate set to $6\mathrm{e}{-5}$. To ensure an effective learning rate, we linearly decay the initial rate at every iteration so at the end of the training is $20\%$ of the initial value. We use a dropout rate of $0.3$ on the score projection and dimensionality reduction neurons. As our model is agnostic to the image input size (as long as it can be divided in $14\times 14$ patches), we evaluate on images of size $322\times322$ but train on $224\times224$ to speedup training time.

To \textbf{validate} our experiments and select the hyperparameters, we monitored the recall in the Pittsburg30k-test~\cite{torii2013visual}. We observed that, in the long run, most configurations perform similarly, but rapid convergence on a few epochs is more sensitive to the hyperparameters.

\subsection{Results}
\label{subsec:results}

We benchmarked our model against several single-stage baselines, namely NetVLAD~\cite{arandjelovic2016netvlad} and GeM~\cite{radenovic2018fine} as two representative tradicional baselines, and Conv-AP~\cite{ali2022gsv}, CosPlace~\cite{berton2022rethinking}, MixVPR~\cite{ali2023mixvpr} and EigenPlaces~\cite{berton2023eigenplaces} as the four most recent and best performing baselines in the literature. The evaluation spanned a diverse array of well-established datasets: MSLS Validation and Challenge~\cite{warburg2020mapillary}, which are comprised of dashcam images; Pittsburgh250k-test~\cite{torii2013visual}, featuring urban scenarios; SPED~\cite{chen2018learning}, a collection from surveillance cameras; NordLand, notable for its seasonal variations from images captured from the front of a train traversing Norway; and SF-XL~\cite{berton2022rethinking}, a large urban dataset to evaluate VPR at scale. We use Recall@k (R@k) as the metric for all our experiments, as it is standard in related work. We use evaluation data and code from MixVPR~\cite{ali2023mixvpr}, which considers retrieval as correct if an image at less than $25$ meters (or two frames for Nordland) from the query is among the top-k predicted candidates.

As shown in Table \ref{tab:main}, our model outperforms all previous methods on all datasets and all metrics. Even the smaller $512+32$ version already surpasses previous models with bigger descriptors on most datasets. It is worth highlighting the metrics saturation observed in MSLS Val, Pitts250k-test and SPED, and on the other hand the challenging nature of MSLS Challenge and NordLand. The MSLS Challenge dataset, with its diversity, extensive size and closed labels, and NordLand, with its extreme sample similarity and seasonal shifts, emerge then as key benchmarks for assessing VPR performance. Although our DINOv2 SALAD shows a significant improvement on \textit{all} benchmarks, it is precisely in MSLS Challenge and NordLand where we obtain the most substantial recall increases, with $+7.6\%,+11.7\%$ and $+17.6\%,+14.6\%$ for R@1, R@5 respectively over the second best. For SF-XL, as shown in Table \ref{tab:sfxl}, our method also achieves the best results to date. This is remarkable, considering that the previous state of the art was trained on this dataset, whereas our method never used any image of San Francisco when it was fine-tuned.

\begin{table*}[ht!]
    \centering
    \resizebox{\linewidth}{!}{
    \begin{tabular}{l c c c c c c c c c c c c c c c c c c c c}
    \hline

    \hline
    \multirow{2}{*}{Method} &  & \multicolumn{3}{c}{MSLS Challenge}&&\multicolumn{3}{c}{MSLS Val}&&\multicolumn{3}{c}{NordLand}&&\multicolumn{3}{c}{Pitts250k-test}&&\multicolumn{3}{c}{SPED}\\
    \cline{3-5} \cline{7-9} \cline{11-13} \cline{15-17} \cline{19-21}
    &  Desc. size & R@1 & R@5 & R@10 && R@1 & R@5 & R@10 && R@1 & R@5 & R@10 && R@1 & R@5 & R@10 && R@1 & R@5 & R@10\\
    \hline
        ResNet NetVLAD~\cite{arandjelovic2016netvlad} & 32768 & 35.1 & 47.4 & 51.7 && 82.6 & 89.6 & 92.0 && 32.6 & 47.1 & 53.3 && 90.5 & 96.2 & 97.4 && 78.7 & 88.3 & 91.4\\
        DINOv2 AnyLoc~\cite{keetha2023anyloc} & 49152 & 42.2 & 53.5 & 58.1 && 68.7 & 78.2 & 81.8 && 16.1 & 25.4 & 30.4 && 87.2 & 94.4 & 96.5 && 85.3 & 94.4 & 95.4\\
        \hline
        ResNet SALAD & 8192 & 57.4 & 70.8 & 74.9 && 83.2 & 89.5 & 91.8 && 33.3 & 49.6 & 55.8 && 91.4 & 96.9 & 97.9 && 75.0 & 86.7 & 89.8\\
        ConvNext~\cite{liu2022convnet} SALAD & 8192 & 63.9 & 75.2 & 80.1 && 85.5 & 92.4 & 94.5 && 47.8 & 64.3 & 70.3 && 93.9 & 97.9 & 98.8 && 83.5 & 90.9 & 92.9\\
        \hline
        DINOv2 GeM & 4096 & 62.6 & 78.3 & 83.0 && 85.4 & 93.9 & 95.0 && 35.4 & 52.5 & 59.6 && 89.5 & 96.5 & 98.0 && 83.0 & 92.1 & 93.9\\
        DINOv2 MixVPR & 4096 & 72.1 & 85.0 & 88.3 && 90.0 & 95.1 & 96.0 && 63.6 & 80.1 & 84.6 && 94.6 & 98.3 & \textbf{99.3} && 89.8 & 94.9 & 96.1\\
        DINOv2 NetVLAD & 24576 & \textbf{75.8} & 86.5 & 89.8 && \textbf{92.4} & 95.9 & 96.9 && 71.8 & 86.5 & 90.1 && \textbf{95.6} & \textbf{98.7} & \textbf{99.3} && 90.8 & 95.7 & \textbf{96.7}\\
        DINOv2 NetVLAD (dim. red.) & 8192 & 73.3 & 85.6 & 88.3 && 90.1 & 95.4 & 96.8 && 70.1 & 86.5 & 90.2 && 95.4 & 98.4 & 99.1 && 90.6 & 95.4 & \textbf{96.7}\\
        \textbf{DINOv2 SALAD (ours)} & 8192 + 256 & 75.0 & \textbf{88.8} & \textbf{91.3} && 92.2 & \textbf{96.4} & \textbf{97.0} && \textbf{76.0} & \textbf{89.2} & \textbf{92.0} && 95.1 & 98.5 & 99.1 && \textbf{92.1} & \textbf{96.2} & 96.5\\
    \hline

    \hline
    \end{tabular}}
    \caption{\textbf{Ablations}. The first two rows correspond to two baselines in the literature~\cite{arandjelovic2016netvlad,keetha2023anyloc}, the rest to different aggregations appended to DINOv2 including our DINOv2 SALAD. Note that only DINO NetVLAD, with a significantly bigger descriptor size than ours, is able to show competitive results. We outperform all the rest DINOv2 baselines of similar descriptor sizes by a large margin.}
\label{tab:avladtions}
\end{table*}

In Table \ref{tab:reranking}, we compare our DINOv2 SALAD method, which solely operates on a single retrieval stage, against the leading two-stage Visual Place Recognition (VPR) techniques. In this comparison, we include the best performing models in the literature, namely R2Former~\cite{zhu2023r2former}, TransVPR~\cite{wang2022transvpr}, and Patch-NetVLAD~\cite{hausler2021patch}, which incorporate a re-ranking refinement. 
Note how our DINOv2 SALAD, despite being orders of magnitude faster and smaller in memory, significantly outperforms all these two-stage methods on all benchmarks. This finding not only highlights the efficiency of our model but also demonstrates the effectiveness of global retrieval using our novel SALAD aggregation. Additionally, considering our method's reliance on local features, we believe that a re-ranking stage could also be applied, potentially increasing our recall metrics but at the price of a higher computational footprint.

\subsection{Ablation Studies}
\label{subsec:ablation}

\textbf{Effect of DINOv2}. 
We assess the impact of the DINOv2 backbone and our optimal transport aggregation SALAD separately. For this, we compare with the existing baselines of ResNet NetVLAD or AnyLoc, this last one applying a VLAD on top of a pretrained DINOv2 encoder.
We integrate the DINOv2 backbone with various aggregation modules, obtaining a handful of performant techniques that improve their respective previous results. As shown in \Cref{tab:avladtions}, all of these outperform the baselines, even though AnyLoc already uses DINOv2. This validates the DINOv2's integration in end-to-end fine-tuning to refine its capabilities.

\textbf{Effect of SALAD}. 
Our experiments in \Cref{tab:avladtions} show that aggregation also matters. Even the recent MixVPR aggregation coupled with DINOv2 does not match the performance of DINOv2 NetVLAD and DINOv2 SALAD. We believe that the DINOv2 backbone is especially suitable for local feature aggregation, as its features work remarkably well in dense visual perception tasks~\cite{oquab2023dinov2, kappeler2023few,yao2023vitmatte}. Although DINOv2 NetVLAD achieves comparable performance to SALAD, it employs a descriptor almost three times as big. Besides, the generalization performance of DINOv2 NetVLAD is limited, as observed in NordLand results. We attribute this to NetVLAD's priors initialization with urban scenarios, which constrain the convergence of the system. In our experiments we also trained a slimmer DINOv2 NetVLAD version, whose features are dimensionally reduced as described in \Cref{subsec:aggregation}, targetting a final descriptor of roughly the same size as SALAD. In this fairer setup, DINOv2 SALAD clearly outperforms DINOv2 NetVLAD. We also evaluate SALAD on top of ResNet and ConvNext backbones, which improves over baseline ResNet NetVLAD but is significantly worse than using DINOv2. This indicates that SALAD is specially suited for high spatial resolution features, like the ones from DINOv2.

\textbf{Effect of hyperparameters}.
DINOv2 comes in different sizes that affect the number of parameters, inference speed, and representation capabilities. As shown in \Cref{tab:dinosizes}, more parameters do not always result in better performance. Excessively big models might be harder to train or prone to overfitting the training set. From these results, we chose the DINOv2-B backbone, which exhibits a great balance between performance and size and speed. Regarding descriptor size, we observed (\Cref{tab:main}) that changing $m$ and $l$ allows to get slimmer versions with competitive performance. For the number of blocks to train, as shown in \Cref{tab:salad_blocks}, fine-tuning two or four block report the best results without significant computation overhead.

\begin{table}
    \centering
    \small
    \resizebox{\linewidth}{!}{
    \begin{tabular}{l c c c c}
    \hline

    \hline
    Model & Dim. size & \# Params. & Latency (ms) & MSLS Val R@1 \\
    \hline
        S & 384 & 21M & 1.30 & 90.5\\
        B & 768 & 86M & 2.41 & 92.2 \\
        L & 1024 & 300M & 7.82 & 92.6 \\
        G & 1536 & 1100M & 24.93 & 91.7\\
    \hline

    \hline
    \end{tabular}
    }
    \caption{\textbf{DINOv2 configurations and performances.}} 
\label{tab:dinosizes}\vspace{-1em}
\end{table}

\begin{figure*}
  \centering
   \includegraphics[width=0.95\linewidth]{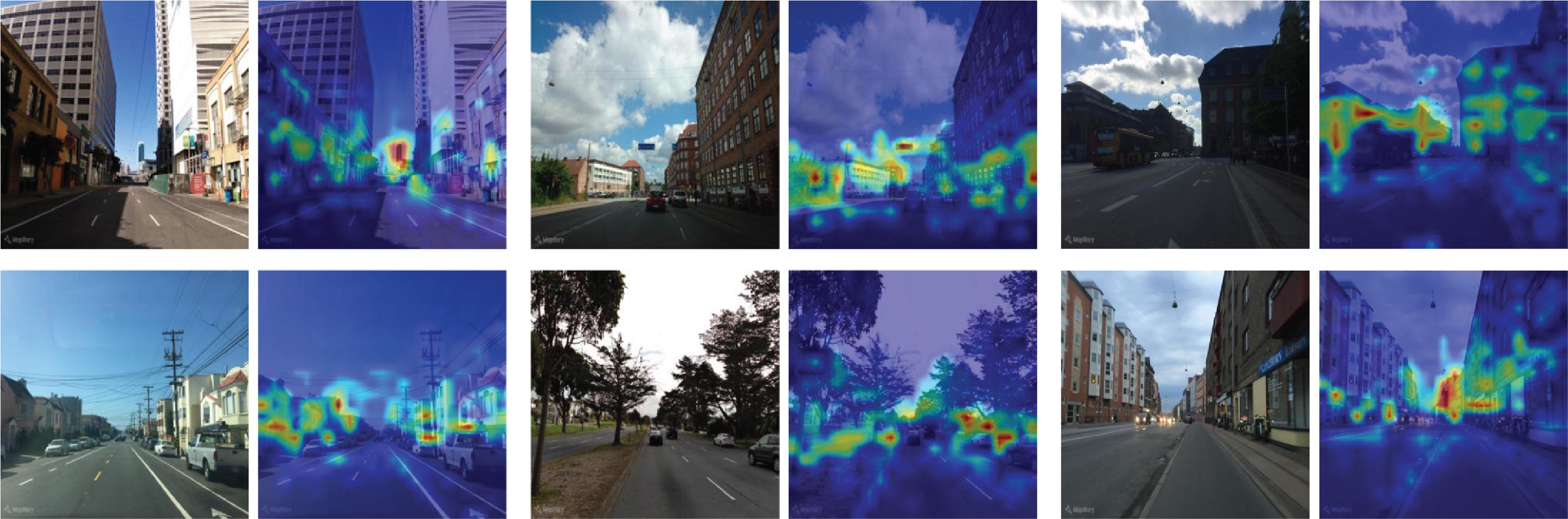}
   \caption{\textbf{Heatmap of local features importance}. Left images show the original pictures, their right counterparts represent the weights \textit{not} assigned to the `dustbin'. Note how the network learns to discard uninformative regions like skies, roads or dynamic objects, and instead focus on distinctive patterns in buildings and vegetation. We attribute its focus on distant buildings to their invariance to viewpoint change.}
   \label{fig:dustbin}
\end{figure*}

\textbf{Effect of SALAD components}.
In \Cref{tab:salad_components}, we show how different components of our SALAD pipeline affect the final performance. Both the global token, which appends global information not captured in local features, and the dustbin, which helps in distilling the aggregated features, contribute to the performance of SALAD. We also trained a model using a dual-softmax~\cite{rocco2018neighbourhood} to solve the optimal transport assignment, following  LoFTR and Gluestick~\cite{sun2021loftr, pautrat2023gluestick}. Although dual-softmax achieves only slightly worse performance, the Sinkhorn Algorithm is theoretically sound and provides a better acronym to our method.

\begin{table}
    \centering
    \resizebox{\linewidth}{!}{
    \begin{tabular}{l c c c}
    \hline

    \hline
    \multirow{2}{*}{Method} & \multicolumn{3}{c}{MSLS Val}\\
    \cline{2-4}
    & R@1 & R@5 & R@10\\
    \hline
        DINOv2 SALAD (frozen) & 88.5 & 95.0 & 96.2 \\
        DINOv2 SALAD (train 2 last blocks) & 92.0 & \textbf{96.5} & \textbf{97.0}\\
        DINOv2 SALAD (train 4 last blocks) & \textbf{92.2} & 96.4 & \textbf{97.0}\\
        DINOv2 SALAD (train 6 last blocks) & 91.6 & 96.2 & \textbf{97.0}\\
        DINOv2 SALAD (train all blocks) & 89.2 & 95.1 & 96.1\\

    \hline
    \end{tabular}}
    \caption{\textbf{Fine-tuning different number of DINOv2 blocks.}}
\label{tab:salad_blocks}
\end{table}\vspace{-0.5em}
\begin{table}
    \centering
    \resizebox{\linewidth}{!}{
    \begin{tabular}{l c c c}
    \hline

    \hline
    \multirow{2}{*}{Method} & \multicolumn{3}{c}{MSLS Val}\\
    \cline{2-4}
    & R@1 & R@5 & R@10\\
    \hline
        DINOv2 SALAD w/o dustbin & 91.4 & 95.8 & 96.2\\
        DINOv2 SALAD w/o global token & 91.8 & 96.0 & 96.2\\
        DINOv2 SALAD (Dual Softmax) & 91.9 & 95.7 & 96.5\\
    \hline
        DINOv2 SALAD & \textbf{92.2} & \textbf{96.4} & \textbf{97.0}\\
    \hline

    \hline
    \end{tabular}}
    \caption{\textbf{Ablation study of the SALAD components.}}
\label{tab:salad_components}
\end{table}

\subsection{Introspective Results}
\label{subsec:qualitative}

\begin{figure*}[!htb]
  \centering
   \includegraphics[width=0.98\linewidth]{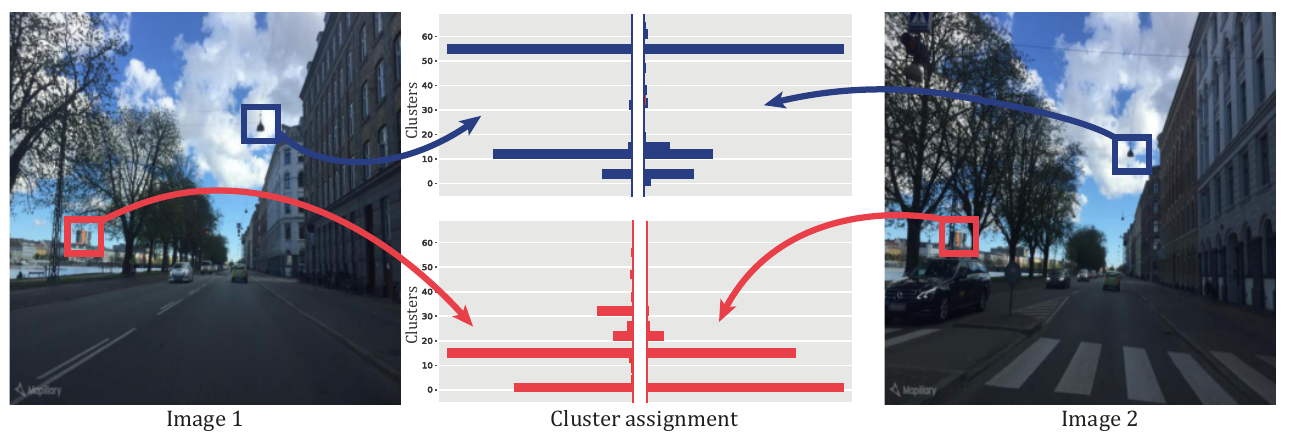}
   \caption{\textbf{Illustration of feature-to-cluster assignments.} See at the leftmost and rightmost part of the figure two different views of the same place. Framed by red and blue squares we highlight two corresponding patches in each of the images. The central part of the figure shows the feature-to-cluster assignments for these patches. Note how DINOv2 SALAD correctly assigns the features to the same bins for both views, even with different local texture. }
   \label{fig:distributions}\vspace{0.5em}
  \centering
   \includegraphics[width=0.98\linewidth]{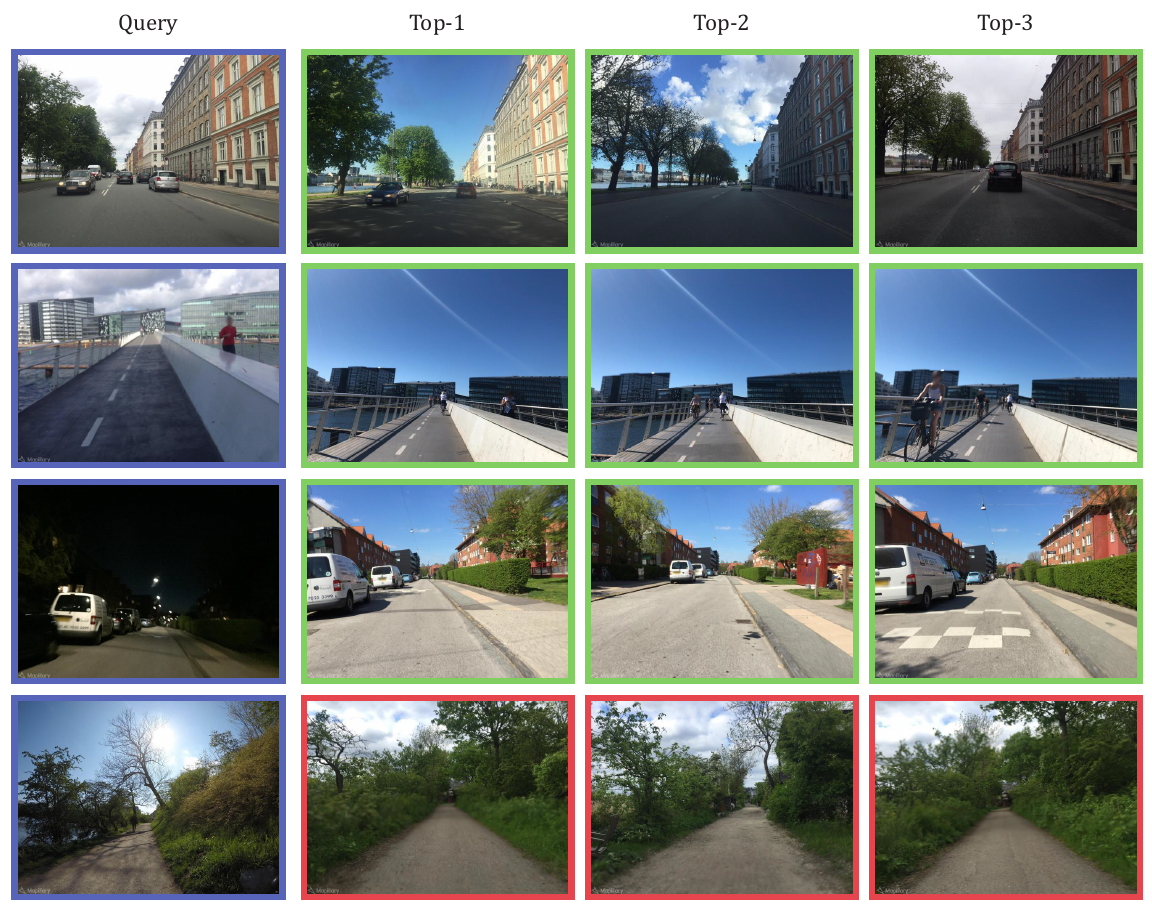}
   \caption{\textbf{DINOv2 SALAD qualitative results at MSLS.} The left column shows several queries and the three other ones shows the top-3 candidates retrieved by our DINOv2 SALAD. Candidates are framed in green if they correspond to the same place as the query, and in red if they do not. Note the correct retrievals under seasonal, weather, viewpoint and day-night changes. Note also a challenging failure case in the last row, due to non-discriminative image content.}
   \label{fig:qualitative}
\end{figure*}

We provide an introspection of our model's performance through a series of illustrative figures. \Cref{fig:dustbin} visualizes the weights that are not assigned to the `dustbin', offering insight into the parts of the input image that the network considers informative. As the `dustbin' assignment is completely learnt by the network, some discarded features might be counter-intuitive. However, we observe that it typically removes dynamic objects and focuses on the most distinctive and invariant parts of the image.
In \Cref{fig:distributions}, we display the assignment distribution of patches from two different images depicting the same place. It demonstrates the model's ability to consistently distribute most of the weights into the same bins for patches representing similar regions. Such repeatable and consistent assignment across different images of the same place is crucial for the reliability and performance of the system.
Finally, in \Cref{fig:qualitative}, we showcase various query images alongside their respective top-3 retrievals made by our system. \mbox{DINOv2 SALAD} is able to retrieve correct predictions even under challenging conditions, such as severe changes in illumination or viewpoint. 

\section{Conclusions and Limitations}

In this paper, we have proposed DINOv2 SALAD, a novel model for VPR that outperforms previous baselines by a substantial margin. This achievement is the result of combining two key contributions: a fine-tuned DINOv2 backbone for enhanced feature extraction and our novel SALAD (Sinkhorn Algorithm for Locally Aggregated Descriptors) module for feature aggregation. Our extensive experiments demonstrate the effectiveness of these modules, highlighting the model's single-stage nature and exceptionally fast training and inference speed.

While our work brings significant improvements in performance, it is not without limitations. Primarily, the adoption of DINOv2 as our backbone results in slower processing speeds compared to ResNet-based methods. Besides, although SALAD is a general aggregation module, its effectiveness is tied to the choice of backbone. It excels with DINOv2, which offers high spatial resolution features, but it is less suited for coarser features. Additionally, in SALAD we use an optimal transport assignment in its simplest form. More sophisticated constraints could improve the resulting assignment, a very relevant aspect for our future work.

\section*{Acknowledgments}
\label{sec:acks}

This work was supported by the Spanish Government (PID2021-127685NB-I00 and TED2021-131150B-I00), the Aragón Government (DGA T45\_23R) and the scholarship FPU20/02342.

\clearpage


{
    \small
    \bibliographystyle{ieeenat_fullname}
    \bibliography{main}
}


\end{document}